\documentclass{article}

\usepackage{arxiv}

\usepackage[utf8]{inputenc} 
\usepackage[T1]{fontenc}    
\usepackage{amssymb}
\usepackage{amsmath}
\usepackage{hyperref}       
\usepackage{url}            
\usepackage{booktabs}       
\usepackage{amsfonts}       
\usepackage{nicefrac}       
\usepackage{microtype}      
\usepackage{cleveref}       
\usepackage{lipsum}         
\usepackage{graphicx}
\usepackage{natbib}
\usepackage{doi}

\title{Diagnosis driven Anomaly Detection for CPS}

\author{
    \href{https://orcid.org/0000-0002-7812-4279}{\includegraphics[scale=0.06]{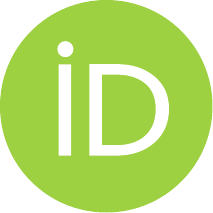}\hspace{1mm}Henrik S.~Steude} \\
    Institute of Automation Technology \\
    Helmut Schmidt University\\
    Hamburg, Germany \\
	\texttt{henrik.steude@hsu-hh.de} \\
	\And
    \href{https://orcid.org/0000-0002-9470-0811}{\includegraphics[scale=0.06]{orcid.pdf}\hspace{1mm}Lukas Moddemann} \\
    Institute of Automation Technology \\
    Helmut Schmidt University\\
    Hamburg, Germany \\
	\texttt{lukas.moddemann@hsu-hh.de} \\
	\And
    \href{https://orcid.org/0000-0002-8674-6895}{\includegraphics[scale=0.06]{orcid.pdf}\hspace{1mm}Alexander Diedrich} \\
    Institute of Automation Technology \\
    Helmut Schmidt University\\
    Hamburg, Germany \\
	\texttt{alexander.diedrich@hsu-hh.de} \\
	\And
    \href{https://orcid.org/0000-0001-5023-839X}{\includegraphics[scale=0.06]{orcid.pdf}\hspace{1mm}Jonas Ehrhardt} \\
    Institute of Automation Technology \\
    Helmut Schmidt University\\
    Hamburg, Germany \\
	\texttt{jonas.ehrhardt@hsu-hh.de} \\
	\And
    \href{https://orcid.org/0000-0001-8747-3596}{\includegraphics[scale=0.06]{orcid.pdf}\hspace{1mm}Oliver Niggemann} \\
    Institute of Automation Technology \\
    Helmut Schmidt University\\
    Hamburg, Germany \\
	\texttt{oliver.niggemann@hsu-hh.de} \\
}

\hypersetup{
pdftitle={Machine Learning, Anomaly Detection,  Diagnosis, Cyber-Physical Systems},
pdfsubject={},
pdfauthor={Henrik S.~Steude, Lukas Moddemann, Alexander Diedrich, Jonas Ehrhardt, Oliver Niggemann},
pdfkeywords={First keyword, Second keyword, More},
}

\begin{document}
\maketitle
\begin{abstract}                
In Cyber-Physical Systems (CPS) research, anomaly detection---detecting abnormal behavior---and diagnosis---identifying the underlying root cause---are often treated as distinct, isolated tasks. 
However, diagnosis algorithms require symptoms, i.e. temporally and spatially isolated anomalies, as input. 
Thus, anomaly detection and diagnosis must be developed together to provide a holistic solution for diagnosis in CPS. 
We therefore propose a method for utilizing deep learning-based anomaly detection to generate inputs for Consistency-Based Diagnosis (CBD). 
We evaluate our approach on a simulated and a real-world CPS dataset, where our model demonstrates strong performance relative to other state-of-the-art models. 
\end{abstract}


\section{Introduction}
\label{sec:intro} Diagnosing system failures, a process that identifies the root causes of malfunctions, is a critical task in many Cyber-Physical Systems (CPS) applications.
The growing complexity of CPS has made it increasingly important to develop diagnostic approaches to ensure their robustness and reliability.
Consistency-Based Diagnosis (CBD) has become the state-of-the-art for complex CPS when limited or no information about possible faults is available \citep{reiter1987theory,DIEDRICH2022104636}.

CBD requires models that represent the normal working behavior of the CPS, typically formulated using propositional logic, comprising symbols for individual components within the system.
Furthermore, CBD needs discrete health states of the system's components, known as observations.
These health states are often generated through anomaly detection methods \citep{jung2018combining,jung2016combined}.
However, diagnosis and anomaly detection are often treated separately in the literature.
Current research in anomaly detection for multivariate time series often employs deep learning methods to identify anomalies at the system level or for individual signals \citep{Garg2022-ol}.
Conversely, the diagnostic literature frequently assumes that labels for anomalies are readily available or can be detected through simple statistical methods \citep{diedrich2021diagnosing}.

Particularly for large and complex CPS, modeling in terms of CBD is a complex task on which the quality of the diagnostic system heavily depends \citep{diedrich2022learning}.
The system can be modeled at different levels within the system hierarchy.
The more detailed the description, the more expensive and error-prone the modeling process becomes.
The ideal level of detail should be aligned with the maintenance strategy \citep{console1999model}.

\begin{figure}[h!]
	\centering
	\includegraphics[scale=0.8]{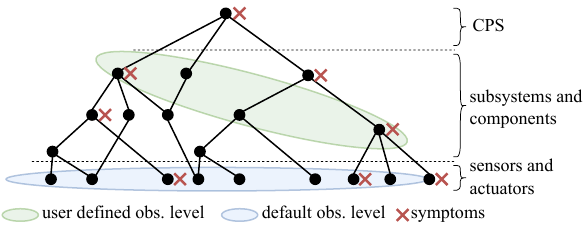}
	\caption{
		Exemplary system overview illustrating subsystems, components, sensors, actuators, and dependencies, with symptoms marked by red crosses.
	}
	\label{fig:system-hierarchy}
\end{figure}

To demonstrate the practical challenges, consider a CPS monitoring thousands of sensor signals.
These signals can be mapped to a substantially smaller set of subsystems, usually less than a hundred.
Having health states available at the subsystem level would greatly simplify the construction of a diagnostic model, such as those proposed by \cite{bunte2019model} or \cite{DIEDRICH2022104636}, as opposed to relying on health states for each sensor.
Yet, as shown in Figure \ref{fig:system-hierarchy}, these aggregated health states are neither readily available nor easy to infer.

In response to this challenge, we introduce an anomaly detection method tailored for identifying anomalies at the subsystem level, aligning with the needs of CBD's system modeling.
This method is specifically designed to manage the complexities of CPS datasets, which are often high-dimensional, time-dependent, and comprise a mix of data types \citep{niggemann2023machine}.

\begin{figure}[h!]
	\centering
	\includegraphics[scale=0.6]{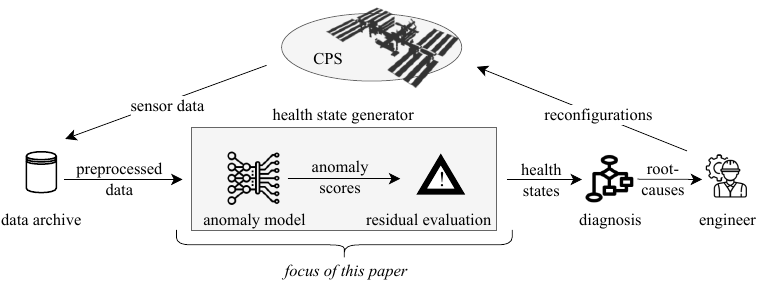}
	\caption{
		Example Use Case of the Columbus Module.
		This figure provides an overview of the high-level information flow in the ISS diagnosis system.
	}
	\label{fig:data-stream}
\end{figure}

The International Space Station's (ISS) Columbus module serves as an example of a complex CPS.
Developing diagnostic software based on telemetry data (see Figure \ref{fig:data-stream}) with over 20,000 signals has motivated the following research questions:
\begin{enumerate}
	\item How effective are state-of-the-art deep learning approaches for anomaly detection at isolating symptoms at an aggregated subsystem level?
	\item How can these methods be optimized for this specific challenge?
\end{enumerate}

In this paper, we assume that the diagnosis model of the CPS, along with the mapping from signals to subsystems, is given as input.
The paper's focus lies on detecting symptoms, rather than on uncovering root causes.
Previous work has shown how to use these health states to identify root causes \cite{DIEDRICH2022104636}.

Given this context, our paper makes two primary contributions:
\textit{(i)} We introduce a novel neural network (NN) architecture specifically designed to generate inputs for CBD algorithms by localizing symptoms within CPS data streams at the subsystem level.
\textit{(ii)} Through a simulated dataset, we highlight the limitations of current models in isolating symptoms at the subsystem level and demonstrate the superior performance of our approach.
Furthermore, using the SWaT dataset \citep{Goh2017-mr}, we show that our model's overall anomaly detection capabilities are competitive with other state-of-the-art models.
The source code for our neural network architecture and the associated experiments is available at our GitHub repository\footnote{https://github.com/hsteude/diag-driven-ad-4-cps.git}.

\section{Related Work}
\label{sec:sota}
Based on the research questions developed in Section \ref{sec:intro}, two research areas are of particular interest to our investigation: diagnostics and anomaly detection for CPS.

The research field of model-based fault diagnosis goes back to the seminal work \cite{reiter1987theory}.
Nowadays, many diagnosis algorithms exist \citep{rodler2022taxonomy}.
Physical system diagnosis has recently been addressed by \cite{muvskardin2020catio} and in the surveys by \cite{dowdeswell2020finding} and \cite{yucesan2021survey}.
\cite{jung2018combining} published an approach to combine Consistency-Based Diagnosis (CBD) with data-driven anomaly classifiers.
Their aim was towards a novel technique for residual generation using one-class support vector machines, whose output was then diagnosed. In another work \cite{lundgren2022data} present a data-driven approach for diagnosis compatible to CBD.
In still other work, \cite{jung2022automated} has used a grey-box approach with neural networks and structural analysis to perform fault diagnosis.
However, none of these approaches integrate anomaly detection with CBD on a logical knowledge base.

In the domain of anomaly detection for multivariate time series, deep learning methods have emerged as the state-of-the-art \citep{Pang2021-fv,Garg2022-ol}.
Broadly, these methods can be divided into reconstruction-based and prediction-based techniques.
Within the group of prediction-based methods, modern network architectures such as Transformers \citep{Chen2022-qa}, TCNs \citep{Cheng2019-sj}, and Graph Neural Networks \citep{Deng2021-vd} are employed.
Within reconstruction-based methods, similar approaches are combined with representation learning techniques like GANs \citep{Li2019-kd,Ciancarelli2023-ax} and VAEs \citep{Chen2022-st,Lin2020-yi}.
Some of the mentioned methods have been explicitly evaluated in the context of CPS data, as demonstrated in \cite{Chen2022-qa} and \cite{Garg2022-ol}.
While some studies address anomaly detection in the context of diagnostics \citep{Marino2021-yv,Garg2022-ol}, none of these works to date have combined deep learning methods for multivariate time series with modern diagnostic techniques.

\section{Solution}
\label{sec:solution}

As outlined in Section \ref{sec:intro}, our objectives are to \textit{(i)} identify symptoms (rather than root causes) in CPS data and \textit{(ii)} localize them at the subsystem level.
This section will formalize the problem and present our proposed solution.

\subsection{Problem Specification}
\label{subsec:problem}

\begin{figure}[h!]
	\centering
	\includegraphics[scale=1.1]{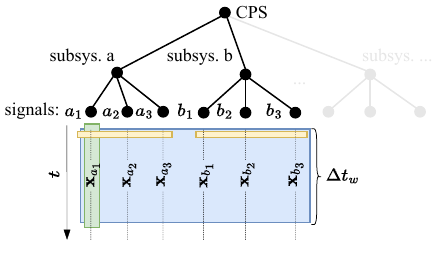}
	\caption{
		Visualization of a simple CPS example. The figure showcases two subsystems \(S=\{a, b\}\), their signals \(\{a_1, ..., b_3\}\), and their time series \(\{\mathbf{x}_{a_1}, ..., \mathbf{x}_{b_3}\}\).
	}
	\label{fig:problem}
\end{figure}

We assume that we have a set of subsystems denoted as \( S \).
Furthermore, we are provided with a mapping, referred to as the \textit{subsystem-signals map}, which associates each subsystem with its corresponding set of CPS signals.
Specifically, for each subsystem \( s \in S \), this mapping defines the subset of signals \( P_s \subseteq P \), where \( P \) is the set of all sensor signals, and \(P_s\) the set of signals associated with subsystem \( s \).
Let \(\mathbf{x}_p\) represent the time series of sensor \(p \in P\), where each entry \(x_p(t) \in \mathbb{R}\) holds the corresponding measurement value at timestamp \(t\).
For a simple example system, these quantaties are visualized in Figure \ref{fig:problem}.

The anomaly detection system, which we call \textit{symptom generator}, can be more formally described as a function \(G\).
This function takes the sensor readings from the most recent time window \(\Delta t_w \in \mathbb{N}\), as input and returns a binary health state for each subsystem.
Let the input \( \mathbf{X}_t \) be the matrix of sensor readings for the \(\Delta t_w\) most recent discrete time steps for all sensors, such that \( \mathbf{X}_t \in \mathbb{R}^{\Delta t_w \times |P|} \).
As output, \(G\) yields a vector \(\mathbf{h}(t)\) where each element \(h_p(t)\) indicates the health state of the \(p\)-th subsystem at time \(t\):
\[
	\mathbf{h}(t) = G(\mathbf{X}_t)
\]
with \( h_p(t) \in \{0,1\} \), where \(0\) signifies "OK" and \(1\) signifies "not OK".
These states can be used to assign values to the \textit{observations} used by \cite{DIEDRICH2022104636}.

\subsection{Proposed NN architecture}
\label{subsec:architecture}

In order to implement \(G\) we introduce a novel NN architecture, which leverages the $\beta$-Variational Autoencoder (VAE) \citep{Higgins2017-aw} framework and adapts the Temporal Convolutional Network (TCN) \citep{Bai2018-kf} for the encoder and decoder NNs. This makes our implementation similar to the TCN-AE introduced by \cite{mengTimeConvolutionalNetwork2020}.
We further integrate prior knowledge of the CPS in the form of a \textit{subsystem-signals map} into the network architecture.
\begin{figure}[h!]
	\centering
	\includegraphics[scale=.7]{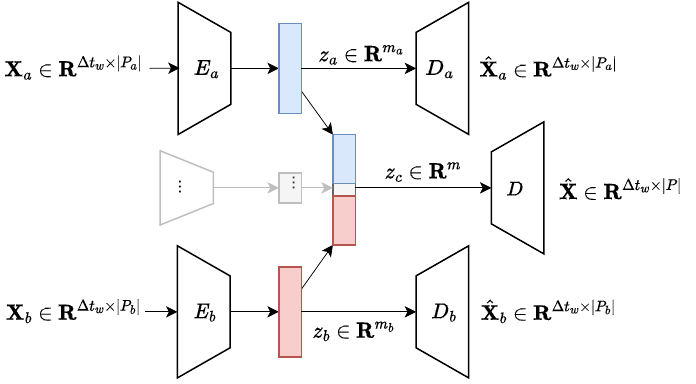}
	\caption{
		High-level model architecture.
		The visualizations of the latent space variables \( \mathbf{z}_a \) and \( \mathbf{z}_b \), output by the encoders for subsystems \( a \) and \( b \), are highlighted in blue and red, respectively.
		The encoders and decoders for additional subsystems are indicated in gray.
	}
	\label{fig:high-level-architecture}
\end{figure}

In the proposed architecture (visualized in Figure \ref{fig:high-level-architecture}), the key innovation is the \textit{composite latent space}.
Signals from distinct subsystems, denoted by \(i \in \{S\}\), are independently processed by dedicated encoder networks \(E_i\).
Each encoder produces a latent representation \(\mathbf{z}_i \in \mathbb{R}^{m_i}\), where \(m_i\) represents the dimensionality of the latent space associated with subsystem \(i\).
These subsystem-specific latent vectors are then concatenated to form the \textit{composite latent space} \(\mathbf{z}_c \in \mathbb{R}^{m}\), satisfying \(m = \sum{m_i}\) and ensuring that \(m\) is significantly less than the product of the time window length \(\Delta t_w\) and the number of signals \(|P|\).
The reconstruction of the full signal set \(P\) is performed by the decoder network \(D\), which takes \(\mathbf{z}_c\) as input.
This architectural design enforces an isolation constraint, preventing inter-subsystem information flow, thereby enhancing the fault isolation capability at the subsystem level.
At the same time, the \textit{composite latent space} \(\mathbf{z}_c\) ensures that cross-subsystem anomalies can be identified.

\begin{figure*}[h!]
	\centering
	\includegraphics[scale=0.6]{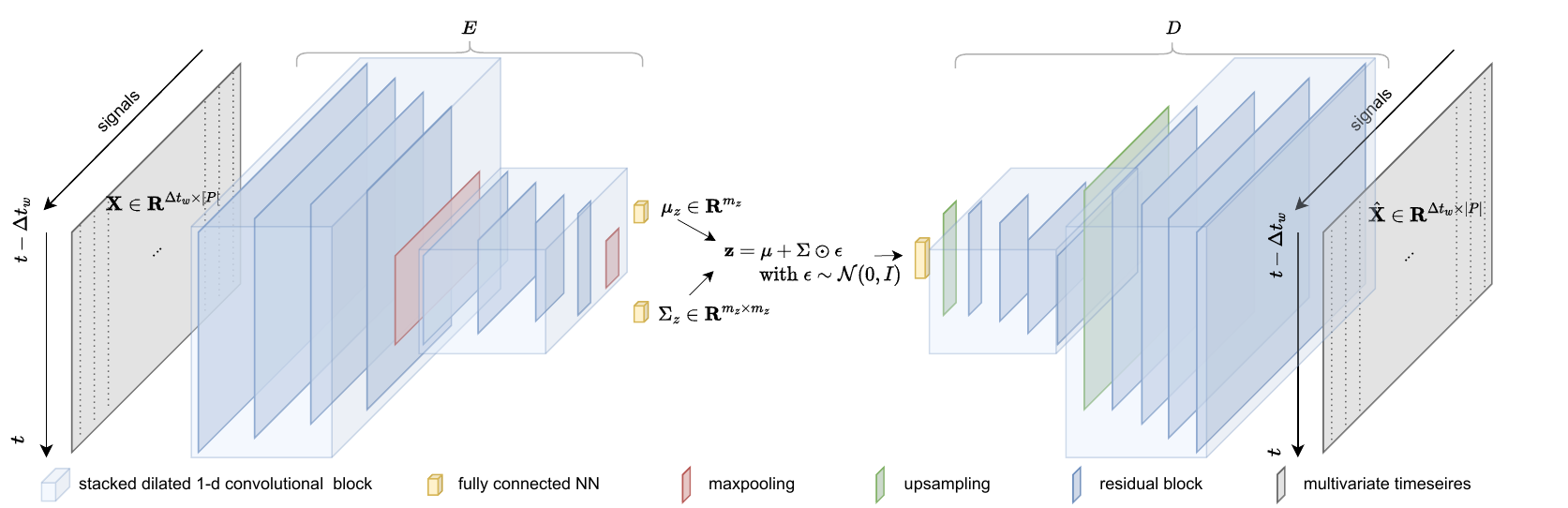}
	\caption{
		Schematic of our TCN-VAE architecture. This architecture is used within \(E_i\) and \(D_{i}\) as referenced in Figure \ref{fig:high-level-architecture}}
	\label{fig:tnc-vae-architecture}
\end{figure*}
The second new component of our architecture is the structure of the individual encoder and decoder NNs.
For simplicity, we describe this structure based on a simple encoder-decoder setup and omit the differentiation between individual subsystems and the corresponding indices.
Figure \ref{fig:tnc-vae-architecture} visualizes the architecture, which is representative of all encoders and decoders of the subsystems.

Our network architecture is built upon \textit{residual blocks}, which consist of two convolutional layers with 'same' padding and equal dilation rates to maintain consistent temporal dimensions as well as skip connections.
These blocks are organized into a \textit{stacked dilated 1-D convolutional structure}, similar to TCNs.
However, instead of using causal convolutions, our design employs non-causal ones, allowing for a full representation of the input sequence.
Within these stacked blocks, channel dimensionality is reduced, followed by a max pooling operation that halves the time series length, thereby achieving temporal dimension reduction.
Ultimately, a fully connected layer is utilized to generate the parameters for the posterior distribution \(q(\mathbf{z|X})\) from the output of the second max pooling layer.
The latent variable \(\mathbf{z}\) sampled from this distribution is fed into the decoder \(D\).
The decoders echo the encoder's architecture, substituting max pooling with upsampling layers to reinstate the original time series dimension.

The loss function we use for adjusting the weights consists of three components: the mean squared error (MSE) for each individual decoder averaged across all decoders, the Kullback-Leibler (KL) divergence averaged across all individual posterior distributions relative to their respective priors, and the global MSE for the entire set of signals.
Formally, the loss function is defined as follows:
\begin{align}
	\begin{split}
		\label{eq:loss}
		\mathcal{L} = & \frac{1}{|S|} \sum_{i=1}^{|S|}\Big[\text{MSE}(\mathbf{X}_i, \mathbf{\hat{X}}_i) + \beta \, \text{KL}\bigl(p(\mathbf{z}_i) \| q(\mathbf{z}_i | \mathbf{X}_i)\bigr)\Big] + \text{MSE}(\mathbf{X}, \mathbf{\hat{X}})
	\end{split}
\end{align}
where each \(\mathbf{X}_i\) holds the signals associated to the subsystem \(i\) in the \textit{subsystem-signals map} and \(\mathbf{\hat{X}}_i\) is the reconstruction i.e. the output of decoder \(i\).
\(\beta \in \mathbb{R}\) is the regularization weight introduced in \cite{Higgins2017-aw}.

\subsection{Residual evaluation}
\label{subsec:residual-eval}
To generate the binary health states \(\mathbf{h}_t\) that our function \(G\) is designed to produce, we use the reconstructions \(\hat{\mathbf{X}}_i\) for all \(i \in S\) as well as \(\hat{\mathbf{X}}\).
This process involves employing various scoring and thresholding methods, as detailed by \citet{Garg2022-ol}, who discuss a range of implementations and their respective pros and cons.
In our implementation, we use the reconstruction error from individual subsystems and the overall error across all signals as our scoring function, and we select the best F1-score method for thresholding, following the approaches presented in the cited work.
Ultimately, the efficacy of the model depends on the difference in reconstruction error between an \textit{OK} sample and a \textit{not OK} sample.
The larger this difference, the easier the binarization.

\section{Experiment}
\label{sec:experiment}
Our main experiment aims to investigate how the \textit{composite-latent-space} performs in comparison to other architectures.
To our knowledge, there is no publicly available CPS dataset that contains symptom labels at the subsystem level, necessitating the use of a simulated dataset for this analysis.
However, we were able to demonstrate the capability of our architecture to detect anomalies within the entire system in a secondary experiment by applying our model to the SWaT dataset \citep{Goh2017-mr}.
With a composite F1 score of 0.52, it ranks among the best of the models benchmarked by \cite{Garg2022-ol}.
We do not further elaborate on this experiment here, as the focus of this paper is on symptom isolation at the subsystem level.
For corresponding details, we refer to our code repository.

\begin{figure}[h!]
	\centering
	\includegraphics[scale=.6]{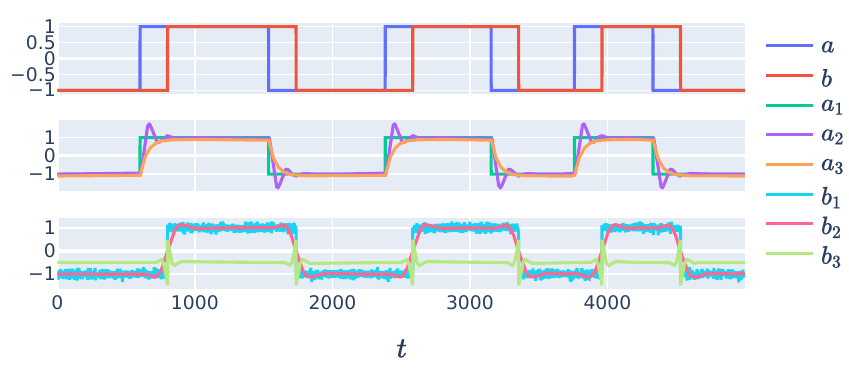}
	\caption{
		Healthy data sample (500 timestamps).
		Causal signals \(a\) and \(b\) (top) with derived signals \(a_1\) to  \(a_3\) and \(b_1\) to \(b_3\) (middle and bottom), used for model training and validation.
	}
	\label{fig:simulation}
\end{figure}

\textbf{Dataset}
The simulation assumes a system structured as shown in Figure \ref{fig:problem}, comprising two subsystems \(S=\{a,b\}\) and a total of six signals \(P=\{a_1, a_2, \ldots, b_3\}\).
The signals \(a\) and \(b\) visualized at the top of Figure \ref{fig:simulation} can be thought of as steering signal for their subsystems.
These causal signals are not included in the dataset, they are only used during data generation.
The length of the high and low states of signal \(a\) is randomly sampled from a uniform distribution, ranging between 500 and 1000 timesteps.
Signal \(b\) is a delayed version of signal \(a\).
Signals \(a_1\), \(a_2\), and \(a_3\) are derived from the causal signal \(a\), with \(a_1\) mirroring \(a\), \(a_2\) representing the response of a second-order dynamic system, and \(a_3\) that of a first-order dynamic system.
Signals \(b_1\), \(b_2\), and \(b_3\) are based on the causal signal \(b\), where \(b_1\) is signal \(b\) with added noise, and \(b_2\) and \(b_3\) are its low-pass and high-pass filtered versions, respectively.

\textbf{Benchmark models}
We have selected four modeling approaches commonly utilized in practice as baselines:
\begin{enumerate}
	\item A \textit{vanilla TCN-VAE}, consisting of an extensive encoder and decoder that processes all signals from the CPS, reconstructing the input data (illustrated in blue in Figure \ref{fig:problem}).
	      For the computation of the reconstruction error each  subsystems, the average error of the signals associated with the subsystem is used.
	\item A \textit{univariate TCN-VAE} whose encoder and decoder also follow the architecture depicted in Figure \ref{fig:tnc-vae-architecture}, but models each signal in \(P\) independently (shown in green in Figure \ref{fig:problem}).
	      The subsystem-specific reconstruction error is also derived by aggregating the errors of the corresponding signals.
	\item A \textit{Gaussian Mixture Model (GMM)} trained on individual time points, which only captures the data distributions at each timestamp.
	      Here, a separate model is trained for each subsystem to derive a specific anomaly score (represented in yellow in Figure \ref{fig:problem}).
\end{enumerate}

To ensure a fair comparison of the models' performance, all models, with the exception of the GMM, share the same TCN-VAE architecture for their encoders and decoders.
Additionally, the total number of parameters for all neural network models was kept in the same order of magnitude, approximately 500k.
The total number of latent variables across all models was uniformly maintained at 12, and each model underwent independent hyperparameter tuning.
With the optimal set of hyperparameters, the final models were trained using early stopping based on the validation loss.

\textbf{Testing dataset}
The simulation described above represents the nominal operational condition, which we divide into training and validation sets.
To assess the anomaly detection capabilities of the proposed model, we introduce four distinct fault scenarios into the dataset:

\begin{itemize}
	\item Fault 1: The signal \(a_1\) remains constant at a value of -1, simulating a stuck-at fault condition.
	\item Fault 2: An offset is introduced to signal \(b_3\), elevating its value by +1, reflecting a calibration or drift fault.
	\item Fault 3: A temporal shift is applied to all signals within subsystem \(b\), mimicking a delay fault.
	\item Fault 4: The signals for both subsystems are modulated to operate at twice their normal frequency, representing a speed or performance anomaly.
\end{itemize}

The test set consists of 100 samples from each of the four distinct fault scenarios (1-4) and 400 samples from the healthy state, ensuring a balanced representation of labels.
We utilized a binary labeling scheme, where \(0\) denotes the absence and \(1\) signifies the presence of faults within the respective subsystems and across the entire signal set, as outlined in Table \ref{tab:fault_label_mapping}.
\begin{table}[h]
	\centering
	\caption{Label allocation in the test set.}
	\begin{tabular}{cccc}
		\hline
		\textbf{Fault type} & \textbf{Subsys. \(a\)} & \textbf{Subsys. \(b\)} & \textbf{All signals} \\ \hline
		Healthy             & 0                      & 0                      & 0                    \\
		Fault 1             & 1                      & 0                      & 1                    \\
		Fault 2             & 0                      & 1                      & 1                    \\
		Fault 3             & 0                      & 0                      & 1                    \\
		Fault 4             & 1                      & 1                      & 1                    \\ \hline
	\end{tabular}
	\label{tab:fault_label_mapping}
\end{table}

Thresholds for anomaly detection in each model and subsystem were computed to optimize their respective F1 scores.
The results are detailed in the subsequent section.

\section{Results}
\label{sec:results}

The performance metrics from our experimental analysis are presented in Table \ref{tab:results}.
Our model demonstrates a consistent improvement in F1 scores across individual subsystems, highlighting its effectiveness in fault detection.
For system-wide symptom identification, our model's performance is comparable to that of the \textit{Vanilla TCN-VAE}, as anticipated.
The development focus of the \textit{composite-latent-space} method was the improvement of symptom isolation at the subsystem level, rather than enhancing detection capabilities across the entire system.

\begin{table}[h!]
\centering
\caption{Evaluation Results of Models Across Different Subsystems.}
\begin{tabular}{llccc}
\hline
\textbf{Model} & \textbf{Subsys.} & \textbf{F1} & \textbf{Precision} & \textbf{Recall} \\
\hline
GMM & a & 0.436 & 0.982 & 0.28 \\
Univar. TCN-VAE & a & 0.662 & 1 & 0.495 \\
Vanilla TCN-VAE & a & 0.854 & 0.82 & 0.89 \\
Our model & a & \textbf{0.945} & 1 & 0.895 \\
\hline
GMM & b & 0.664 & 0.99 & 0.5 \\
Univar. TCN-VAE & b & 0.973 & 0.952 & 0.995 \\
Vanilla TCN-VAE & b & 0.772 & 0.636 & 0.98 \\
    Our model & b & \textbf{0.998} & 0.995 & 1 \\
\hline
GMM & all & 0.811 & 1 & 0.682 \\
Univar. TCN-VAE & all & 0.675 & 0.516 & 0.978 \\
Vanilla-TCN & all & 0.941 & 0.989 & 0.898 \\
Our model & all & \textbf{0.948} & 0.997 & 0.902 \\
\hline
\end{tabular}
\label{tab:results}
\end{table}

\begin{figure}[h!]
	\centering
    \includegraphics[scale=.4]{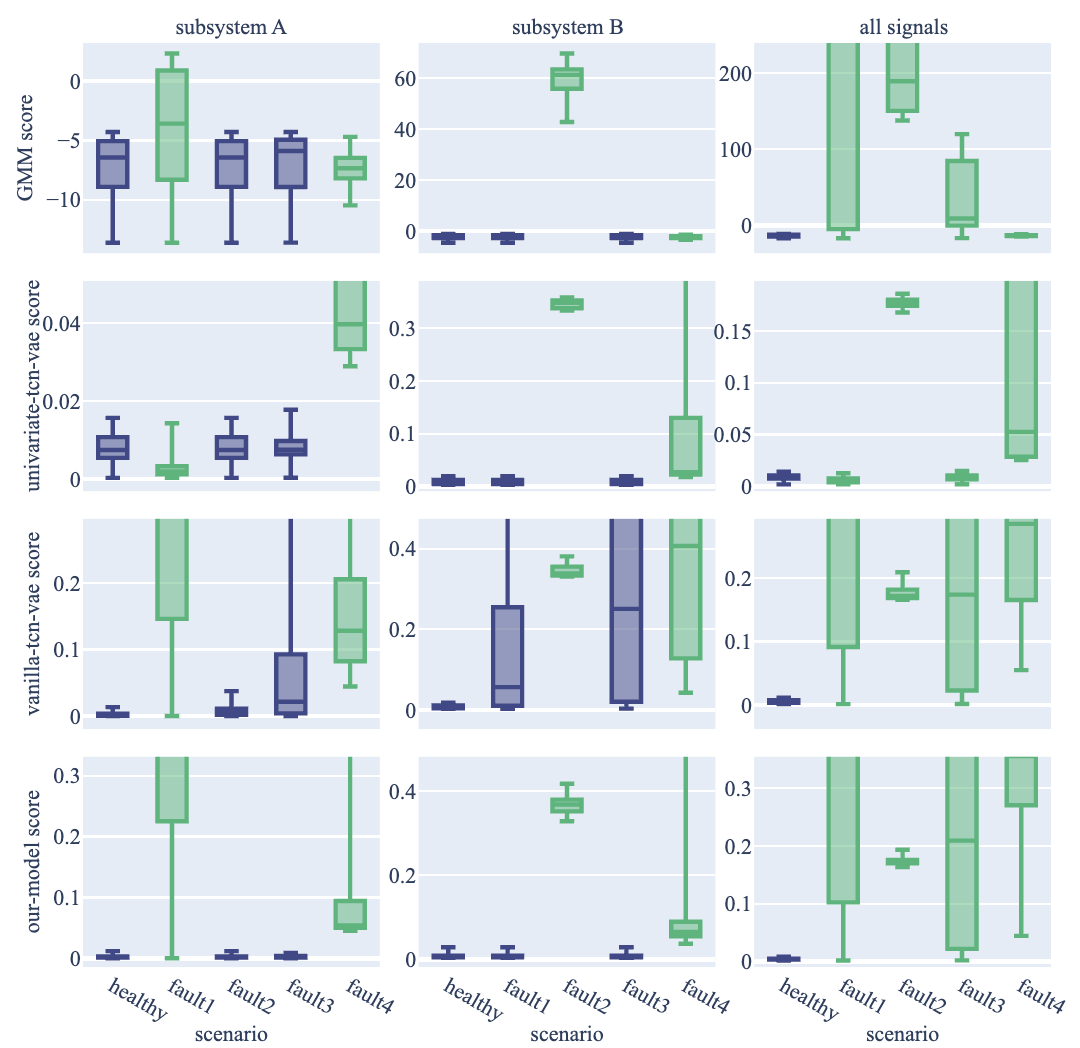}
    \caption{Distributions of model scores across different fault scenarios for subsystems \(a\), \(b\), and the aggregation of all signals. The green color indicates the presence of a symptom in the respective subsystem, while the blue color signifies the absence of a symptom.}
	\label{fig:box}
\end{figure}

For a more detailed view of the results, Figure \ref{fig:box} visualizes the distributions of reconstruction errors—or negative log likelihoods in the case of the GMM—for each model and fault scenario.

The rightmost column of the plot indicates that only the Vanilla TCN-VAE and our proposed model consistently differentiate most fault conditions from the healthy state.

The GMM does not exhibit a significant difference in the score distributions of fault 4 and the healthy scenarios.
Similarly, the univariate model does not effectively distinguish between fault 1 and the healthy state.
The GMM's inability to recognize frequency changes in fault 4 stems from its focus on individual time points rather than on temporal patterns.
Likewise, the univariate model lacks the capacity to detect temporal shifts between signals, as exemplified by fault 3.

However, the most notable observation concerns the Vanilla TCN-VAE model's lack of precision within subsystems \(a\) and \(b\).
The reconstruction error distributions demonstrate that a symptom in subsystem \(a\) significantly increases the reconstruction error in subsystem \(b\), and vice versa (faults 1 and 2).

This suggests that in standard reconstruction-based latent space models, the location of the greatest reconstruction error does not necessarily indicate the actual location of the symptom. The findings highlight that the \textit{composite-latent-space} model effectively remedies this limitation.

\section{Conclusion}
\label{sec:conclusion}

In this study, we introduced a novel neural network architecture, termed the \textit{composite-latent-space} architecture, designed to enhance symptom isolation capabilities within CPS at a user-defined level in the system hierarchy.
The model was evaluated using a simulated dataset designed to represent various fault conditions.

Our results demonstrate that our model surpasses traditional models in isolating symptoms to specific subsystems.
It is particularly interesting to note that larger holistic deep learning models, such as a sequence-to-sequence variational autoencoder without a specialized latent space structure, exhibited weak precision at the subsystem level.

While our model’s performance on the SWaT dataset illustrates its capability for system-wide anomaly detection, our focus is on the model’s superior symptom isolation at the subsystem level.
This specificity in symptom identification is critical in the context of CBD, where binary health states for each subsystem are required as inputs.

For future work, we plan to evaluate the complete approach by diagnosing a comprehensive example system using subsystem health states generated by our proposed method.
This will not only demonstrate the diagnostic capability at the macro level but also validate the practical applicability and effectiveness of our model in real-world CBD use cases.

\bibliographystyle{unsrtnat}
\bibliography{references}  






\end{document}